# From exemplar to copy: the scribal appropriation of a Hadewijch manuscript computationally explored

Wouter Haverals[*], Mike Kestemont

University of Antwerp, Belgium

*Corresponding author: Wouter Haverals wouter.haverals@uantwerpen.be

**Abstract**
This study is devoted to two of the oldest known manuscripts in which the oeuvre of the medieval mystical author Hadewijch has been preserved: Brussels, KBR, 2879-2880 (ms. A) and Brussels, KBR, 2877-2878 (ms. B). On the basis of codicological and contextual arguments, it is assumed that the scribe who produced B used A as an exemplar. While the similarities in both layout and content between the two manuscripts are striking, the present article seeks to identify the differences. After all, regardless of the intention to produce a copy that closely follows the exemplar, subtle linguistic variation is apparent. Divergences relate to spelling conventions, but also to the way in which words are abbreviated (and the extent to which abbreviations occur). The present study investigates the spelling profiles of the scribes who produced mss. A and B in a computational way. In the first part of this study, we will present both manuscripts in more detail, after which we will consider prior research carried out on scribal profiling. The current study both builds and expands on Kestemont (2015). Next, we outline the methodology used to analyse and measure the degree of scribal appropriation that took place when ms. B was copied off the exemplar ms. A. After this, we will discuss the results obtained, focusing on the scribal variation that can be found both at the level of individual words and n-grams. To this end, we use machine learning to identify the most distinctive features that separate manuscript A from B. Finally, we look at possible diachronic trends in the appropriation by B's scribe of his exemplar. We argue that scribal takeovers in the exemplar impacts the practice of the copying scribe, while transitions to a different content matter cause little to no effect.

# INTRODUCTION

Among the Royal Library of Belgium's (KBR) extraordinarily rich collection are two fourteenth-century manuscripts that are of great importance to the field of medieval Dutch literature in general, and that of mysticism in the Low Countries in particular: KBR 2879-80 and KBR 2877-78. Both manuscripts contain the complete oeuvre – consisting of letters, visions, songs, and poems – of the mystical writer Hadewijch.[1]

---

[1] Since unambiguous biographical data are lacking, the historical figure of Hadewijch is largely shrouded in mystery. Through her work, however, one can get a modest glimpse of who she was and when she lived. Researchers who undertook this quest situate Hadewijch in the religious women's movement (*mulieres religiosae*) of the thirteenth century [Mommaers, 2003; Fraeters & Willaert, 2009, p. 13-19; Fraeters, 2013; Willaert, 2013]. On the basis of a seventeenth-century inscription, we might even pinpoint her to Antwerp, but this claim remains unproven [van Mierlo, 1927]. Undeniable, however, is the influence Hadewijch exerted on vernacular mystical writing, most notably on the work of John of Ruusbroec [Axters, 1964; Reynaert, 1981; Arblaster & Faesen, 2014, p. 21]. For a translation of Hadewijch's work into English, see [Hart, 1980].

As their sigla suggest, the manuscripts can be found in close proximity to one another. Their side-by-side shelving is not coincidental. Generally, it is assumed that KBR 2879-80 served as the exemplar of KBR 2877-78. For this reason, the manuscripts are also referred to as ms. A and ms. B, respectively. As the first researcher to examine the connection between the two manuscripts, father Jozef van Mierlo leaves no room for doubt: "The relationship of manuscripts A and B can be easily determined: B is a copy of A" [van Mierlo, 1924-1925, vol. II, p. 11, my translation]. Evidence offered in support of his claim can be found – as is traditionally the case in manuscript studies – in both the presence and absence of variants. When A contains an error, the reading is similarly corrupt in B. Moreover, Van Mierlo points to the compulsion of the scribe responsible for B to amend the text when he deems it necessary. Usually, an erroneous passage was first adopted from A, before erasing it and replacing it with a correct reading in B [ibid., p. 11-12].

The present study aims to investigate the spelling profiles of the scribes who produced A and B in a computational way. In the first part of this study, we will present both manuscripts in more detail, after which we will consider prior research carried out on scribal profiling. Next, we outline the methodology used to analyse and measure the degree of scribal appropriation that took place when ms. B was copied off the exemplar, ms. A. After this, we will discuss the results obtained, focusing on the scribal variation that is found both at the level of individual words and character sequences. Specifically, we seek an answer to the question of how B's scribe appropriated his exemplar, A. How orthographically similar are these manuscripts? What elements were changed, and in what parts of the work does change occur?

**I HADEWIJCH MANUSCRIPTS A AND B**

The origin of both manuscripts can be situated in the fourteenth century. Based on paleographic observations, A can be traced back more specifically to the second quarter of the fourteenth century [Kwakkel, 1999, p. 29-30]. Since B is a copy of A, it is naturally younger; an origin around 1380 is commonly accepted [ibid., p. 32-33]. As far as the scribal activity is concerned, the authoritative view states that manuscript B is the work of a single scribe [ibid., p. 32]. For manuscript A, the situation is somewhat more complex. For a long time, manuscript A was considered to be the work of two scribes. And indeed, on palaeographical grounds it can be demonstrated that a scribal takeover occurs on fol. 70v [ibid., p. 29]. However, it was long overlooked that another, third scribe was involved in the production of A as well.[2] As his share is rather limited, this scribe was not easily detected. In a prior computational analysis by Kestemont on the Hadewijch manuscripts, it was demonstrated that the spelling profile observed in the first 11 folia differs from that found in the subsequent folia [Kestemont 2015, p. 170-172]. Consequently, since Kestemont's study, manuscript A is considered to be a product of three scribes. The first scribe was responsible for copying the text on folio 1r up until 11va. A second scribe starts copying on 11va, line 5, and continues with it up until fol. 70vb, line 6. Here, finally, a third scribe enters and completes the work on fol. 101v [Kwakkel, 1999, p. 39; Kestemont 2015, p. 170-172].[3]

---

[2] In 1925, Van Mierlo had already tentatively hinted at the possibility of a scribal takeover in Brussels, KBR, 2879-2880, fol. 11v [van Mierlo, 1924-1925, vol. II, p. 5]. However, his hypothesis was not adopted in further research. Only in 2015, Kestemont was able to demonstrate that three scribes indeed collaborated on manuscript A [Kestemont, 2015].

In terms of content, manuscripts A and B are virtually identical. They both start off with Hadewijch's prose letters, then proceed to the visions followed by a list of 'perfect' people, next come the songs, and finally the mixed poems. The mirroring is disrupted near the end. Whereas manuscript A ends after the mixed poems, B continues with the so-called *Mengeldichten* 17-29 ('mixed poems'), and a small tractate, entitled *Tweevormich tractaetken*. Both in form and in content, however, these two writings are far removed from Hadewijch's original work [van Mierlo, 1952, p. xxvii-xxxiii and 185-186; Ruh, 1992, p. 331]. Consequently, scholars generally agree that these should not be considered original work by Hadewijch. Rather, they are attributed to pseudo-Hadewijch.[4] A codicological caesura in B moreover renders it plausible that the work of pseudo-Hadewijch was added to the work of Hadewijch at a later stage [Kwakkel, 1999, p. 33].

It should be noted that next to A and B, a third manuscript containing the complete works of Hadewijch is kept at the University Library (UB) of Ghent (ms. 914). Also referred to as ms. C, the creation of this manuscript is likely situated in the final quarter of the fourteenth century [ibid., p. 34-35]. There are, however, some notable differences between mss. A and B on the one hand, and ms. C on the other. While A and B both start with Hadewijch's letters, C begins with her visionary works (Table 1). Also, the order of the songs slightly differs in C [van Mierlo, 1942, vol. I, p. 132; Fraeters & Willaert, 2009, p. 59]. Similar to B, though, C also attests to the thematic union between the work of Hadewijch and that of pseudo-Hadewijch. Given that there are no codicological caesurae in C, the intertwining of both oeuvres is more intimate in this manuscript than in B [Kwakkel, 1999, p. 37-39]. The order in which Hadewijch's work is presented might seem like a detail, yet it prompts the suggestion of a changing literary transmission: toward the end of the fourteenth century, the different building blocks of Hadewijch's oeuvre could shift; additionally, we find that her work is more naturally merged with that of pseudo-Hadewijch [ibid.].[5]

---

[3] In both ms. A and B, several other scribal hands can be identified, making corrections or adding short (Latin) notes in the margins [Kwakkel, 2002, p. 218-221]. Since their contributions to the main texts are negligible, we will not consider them any further.

[4] The epithet 'pseudo-Hadewijch' is somewhat of a misnomer, given that there is little resemblance to Hadewijch's work.

[5] For the sake of completeness, we should also mention that several other manuscripts exist that contain (sometimes extensive, sometimes very limited) parts of Hadewijch's and/or pseudo-Hadewijch's oeuvre. The sigla of these manuscripts are: Antwerp, Ruusbroecgenootschap, Neerl. 385 II; Brussels, KBR, 2412-13; ibid. 3037; ibid. 3093-95; ibid.; The Hague, Royal Library of the Netherlands, 70 E 5; ibid. 133 H 21; Paris, Bibliothèque Mazarine, 920. For a description of their content, see: [Deschamps, 1970, p. 84].

|  | Brussels, KBR, 2879-80 Ms. A (ca. 1325-1350) | Brussels, KBR, 2877-78 Ms. B (ca. 1380) | Ghent, UB, 914 Ms. C (ca. 1375-1400) |
|---|---|---|---|
| Hadewijch | Prose letters<br>Visions<br>*Lijst der volmaakten*<br>Songs<br>Mixed poems 1-16 | Prose letters<br>Visions<br>*Lijst der volmaakten*<br>Songs<br>Mixed poems 1-16 | Visions<br>*Lijst der volmaakten*<br>Prose letters<br>Songs<br>Mixed poems 1-16 |
| Pseudo-Hadewijch |  | *Tweevormich tractaetken*<br>Mixed poems 17-29 | *Tweevormich tractaetken*<br>Mixed poems 17-29 |

Table 1. Overview of the composition of the three manuscripts – A, B, and C – containing the complete oeuvre of Hadewijch. Internally, the different parts are similar to each other. Only the order of the songs differs in manuscripts A and B compared to manuscript C (song 8 in ms. C comes after song 19 in mss. A and B).

Judging solely on the basis of the manuscripts' contents (i.e. their composition, and the occurrence of variants), we can already make a convincing claim for A and B to be closely related. In addition to the above-mentioned arguments, there is also contextual evidence that supports their kinship. *Ex libris* inscriptions in the manuscripts reveal that around 1400 they both belonged to the book collection of Rooklooster, a priory close to Brussels [Kwakkel, 2002, p. 21].[6] Also from Rooklooster, a book list survived (Brussels, KBR 1351-73, f. 1v), which catalogues all the books in the Dutch vernacular that – supposedly – belonged to the priory around 1395 [De Vreese 1962, p. 62; Lievens, 1971, p. 158]. On this list we find no less than four works by Hadewijch, three of which, according to the inventory itself, begin with the words: "God die de clare minne".[7] Not coincidentally, these are the exact words with which manuscripts A and B begin. The fourth Hadewijch-manuscript mentioned on the inventory would start with a variant of the same phrase ("God die clare minne"). Since manuscript C begins with Hadewijch's visions – and thus an entirely different phrase –, it does not correspond to any of the listed manuscripts.

The fact that both A and B belonged to the priory of Rooklooster's book collection around 1400 does not necessarily mean that the manuscripts were also produced there. Erik Kwakkel [2002] came to this conclusion in his seminal dissertation, which takes as its starting point the aforementioned medieval book list. With great persuasion, Kwakkel can demonstrate that most of the vernacular Dutch books surviving from Rooklooster were actually produced at the nearby carthusian monastery of Herne. This is certainly the case, Kwakkel argues, for manuscript B. The carthusian who copied the text used a small sign for marking corrections – closely resembling a small delta (δ) – that Kwakkel only finds in the

---

[6] The inscriptions read as follows: 'Dit boec es der broed*ere* va*n* s*int* pauwels i*n* zoninghe*n* gheheeten de roede cluse (ms. A, f. 101v); 'Dit boec es der broed*ere* van S*int* pauwels in zonien gheheete*n* te rooden dale' (ms. B, f. 166r).

[7] In Middle Dutch, the listing goes as follows: '¶ Jte*m* noch ~~een boec va~~ drie boeke va*n* hadewighe*n* die beghinne*n* aldus. God die de clare mi*n*ne'. A fourth work by Hadewijch is mentioned near the end of the list: '¶ Jte*m* een boec beghint God die cla-[...]' (Brussels, KBR 1351-73, f. 1v). An ink stain prevents further reading. However, the few words of the incipit that are readable make it clear that this is the beginning of the prose letter collection by Hadewijch.

Low Countries in manuscripts produced in this community [Kwakkel, 2002, p. 108-118]. The sign is omnipresent in B. Manuscript A, however, does not contain this correction-sign, and is therefore generally not attributed to the carthusians of Herne, but rather to the workshop of Godevaert de Bloc, a Brussels-based bookbinder and vendor [Verheyden, 1937, p. 130-132]. Most likely, Godevaert approached the Herne carthusians with the request to produce a copy of his Hadewijch-manuscript [Kwakkel, 2002, p. 151-152]. Such an engagement is not inconceivable: after all, the monks did make frequent connections with the world outside their community. They produced manuscripts not only for themselves, but also for other religious houses, and even for private citizens in the nearby city of Brussels [ibid., p. 137-156].

## II SCRIBAL PROFILING

In what follows, we set out to examine the scribal practices in Hadewijch-manuscripts A and B. In doing so, it is important to recognize that two competing forces are at work. On the one hand, together with Van Mierlo and Kwakkel, we can observe that a faithful copy of A was pursued in the manufacture of B. Based purely on layout features, this already becomes clear:

> "The craftsmanship with which the inhabitants [of the charterhouse of Herne] copied can be witnessed in Brussels, KBR, 2877-78 [ms. B], containing the works of Hadewijch: the formal characteristics of its exemplar, Brussels, KBR 2879-80 [ms. A], were imitated with *such* precision that, except for the handwriting, the two look virtually identical. Both manuscripts have the same layout, and the writing takes up (almost) the same space [...]" [Kwakkel, 2002, p. 119, my translation].

On the other hand, we must also recognize that a medieval manuscript will never be *exactly* the same as its exemplar. No matter how focused or dedicated the scribe, orthographic variation will inevitably sneak in [Driscoll, 2010, p. 89]. These differences can be of various sorts. Depending on the research purpose, one type of variation will be considered significant, while another kind may be trivial [Andrews & Mace, 2013, p. 516-518]. For our study, it is not essential to elaborate on the different possible kinds of variation. This is because we apply a rather rigid condition: when two characters differ from each other, we consider it as variation.

By way of illustration, a brief example demonstrates what to expect in terms of variation. Table 2 shows two parallel passages from A and B with their differences marked. A first difference is found in the spelling of the second-person personal pronoun: where manuscript A addresses the audience with 'u', manuscript B does so with 'v'. The second difference – A: 'waernē', B: 'warnē' – is similarly one of non-essential spelling variation. A large group of variants concerns abbreviations. In the passage at hand, we notice, for example, that A uses a superscript letter 't' to represent 'eit', while in B this suffix is written in full. Finally, we can spot various abbreviated words using a straight macron above a letter (¯), or an apostrophe (') to represent omitted letters.

| | |
|---|---|
| NV willic u waernē eens dincs<br>daer vele scadē ane legh₃<br>Dat segghic u dat dit nu es<br>ene de siecste siech˙ die ōder alt volc<br>es van al dē siecheidē die daer ōd' sijn<br>diere nochtā vele sijn ou' al<br><br>A - 6ra, line 11-18 | NV willic v warnē eens dincs<br>d' vele scadē ane legh₃<br>dat segghic v dat dit nv es<br>ene die siecste ziecheit die ond' al tfolc<br>es vā allē dē ziecheidē die d' ond' sijn<br>diere nochtā vele zij ou' al<br><br>B - 7vb, line 18-27 |

Table 2. Comparison of the orthographies of two short, parallel passages of mss. A and B.

Influential historical linguist Angus McIntosh has defined the cornerstones of modern scribal profiling [McIntosh, 1974; 1975]. The main thesis McIntosh puts forward in his pioneering work states that every medieval scribe has a unique scribal fingerprint, which can be identified both linguistically and paleographically [1975, p. 219-222]. For our study, we are exclusively interested in linguistic characteristics. According to McIntosh [1973, p. 61], there are three possible ways in which a scribe can interact with his exemplar.[8] (1) A scribe may strive to produce an exact transcription of the exemplar. Though, such *literatim* copying occurs only rarely. (2) A scribe can make the exemplar's text fit his language and spelling preferences as much as possible. (3) Finally, a scribe can occupy a position between the above-mentioned strategies, copying the text of his exemplar in some parts conservatively, and then again more liberally.

Based purely on the sample text in Table 2, it is difficult to infer the idiosyncratic linguistic characteristics of the scribes who penned down the text. Rather, the undertaking of scribal profiling should be carried out on a larger scale, taking the entire text into account. Such a large scale that is desirable for scribal profiling is also advocated by McIntosh. At the time he conducted his research, however, this implied a dauntingly large amount of manual labour to be executed (i.e. making diplomatic transcriptions of medieval texts, deciding which features to analyse, marking – and subsequently counting – those features, etc.) [McIntosh 1974, p. 611; McIntosh 1975, p. 220]. In short, in the 1970s there were still quite a few practical hindrances that rendered McIntosh's proposal a Herculean task.[9] The digital revolution has dismantled these obstacles to a great extent. Today, systematic research on scribal profiles is facilitated mainly by two major developments. For one, the computer allows us to automatically extract lists of scribal features alongside their frequencies with relative ease (more on this later). Doing so, it is no longer necessary to limit oneself to a pre-defined – and thus finite – set of scribal features.[10] Secondly, Handwritten Text Recognition (HTR) allows us to assemble large corpora of medieval texts. Tools such as *Transkribus* [Kahle et al., 2017; Muehlberger et al., 2019] and *eScriptorium* [Kiessling et al., 2019] push editorial scholarship nowadays to new, unprecedented levels [see a.o. Alcorn et al., 2019].

---

[8] McIntosh suggests this classification in the context of a dialectological study. However, his phrasing is very general, rendering the presented subdivision also applicable to other, non-dialectological scribal behaviour (e.g. idiosyncratic scribal spelling preferences, the use of particular abbreviations, etc.). See a.o. [Black, 1998; Hiltunen & Peikola, 2007; Carrillo-Linares, 2020].

[9] Nevertheless, McIntosh and his colleagues Samuels and Benskin rose to the task. The result of their enterprise is the highly acclaimed *Linguistic Atlas of Late Medieval English* (LALME).

[10] Deciding on a set of characterising scribal traits is something that is given considerable attention in McIntosh's research. See, for example: [McIntosh 1975, p. 220].

**III MATERIALS**

Similar to the dialectological research of McIntosh and his successors, we use diplomatic transcriptions for our subsequent analyses [McIntosh 1974, p. 602].[11] Compared to the prior computational study on the Hadewijch-manuscripts by Kestemont [2015], there are two notable differences relating both to the size of the corpus under scrutiny, and to its encoding.

Firstly, while Kestemont's study is limited to Hadewijch's letters, for our analyses, we will compare *all* parts that manuscripts A and B have in common. These sections include the letters, the visions, the *Lijst der volmaakten*, the songs, and – finally – the mixed poems 1-16.[12] The pseudo-Hadewijchiana that are part of B are left out of our analyses, because of the simple reason that they are not contained in A. As mentioned above, both of A and B diplomatic transcriptions were used. In part, these transcriptions were obtained through the application of the technique of Handwritten Text Recognition (HTR). Two distinct HTR-models were trained, one for each manuscript.[13] The training was performed on the basis of a representative sample of manually transcribed folia. For A, a model was trained on 86 out of the 202 folia containing text (corresponding to 33,083 words). To avoid an imbalance, we ensured that text by each of the three scribes found in ms. A is present in the training material. Similarly, for B a model was trained on 107 out of 293 folia (28,547 words).[14] After training, the performance of both models was evaluated on the validation material, a held-out set of manually transcribed folia.[15] Because this material has not been encountered by the computer during training, the evaluation of the model on the validation set serves as a solid benchmark. As is common, the performance of an HTR-model is expressed in terms of the percentage of characters the model incorrectly predicts. The model trained for manuscript A achieves a Character Error Rate (CER) of 2.52% on the validation set. For manuscript B, the error rate is slightly higher, being 2.81%. Subsequently, both models were deployed to automatically generate transcriptions for the remaining folia of A and B. We are aware that, because of the HTR pipeline, our analyses in the following sections of this article will be carried out on textual material that inevitably contains transcription errors. Nevertheless, we believe that the obtained CERs of ~2.5% are sufficiently low for linguistic analysis. In stylometric research, the quantitative study of authorial style, it has been demonstrated that one would need to replace or remove at least 60% of the characters in a text in order to effectively destroy the authorial signal (this is the case when character n-grams are used as features, which we will do in our experiments) [Eder, 2013, p. 609].

---

[11] All materials and code used for our computational analyses can be accessed online: https://github.com/WHaverals/hadewijch-scribes
[12] Letters (A: f. 1r-41v; B: 1r-59r), visions (A: 42r-59v; B: 59v-89v), *Lijst der volmaakten* (A: f. 59v-61v; B: 89v-93r), songs (A: f. 62r-87r; B: 93r-132r), mixed poems 1-16 (A: f. 87v-101v; B: 132r-147r).
[13] Both the manual transcription, and the training process of the HTR-models were achieved with the program *Transkribus* (v.1.21.0) [Kahle et al., 2017]. The algorithm used for training is CITlab HTR+. Training was conducted over a course of 50 epochs.
[14] The pseudo-Hadewijchiana were not included in the process of training, nor in that of validation.
[15] The validation set comprises 10% of the total amount of manually transcribed, ground-truth pages of each manuscript.

A second point in which the present study differs from Kestemont's [2015] relates to the encoding of the texts. While the diplomatic editions used by Kestemont were encoded following the guidelines of a dedicated TEI-XML scheme (developed by [Boot & Brinkman, n.d.], the input text for our analyses will be Unicode characters. The advantage is that the same level of detail is maintained, without having to deal with complicated mark-up. Moreover, the text remains human-readable, as the example in Table 2 illustrates. For the rendering of specialised, typically medieval characters, we resort to character encodings proposed by the *Medieval Unicode Font Initiative* (MUFI).[16] Finally, it should be noted that researchers often make further distinctions as to the degree of detail with which a diplomatic transcription is pursued. In this article, we conceptualise our diplomatic transcriptions as 'graphemic reproductions', meaning that the spelling is retained as it appears in the manuscripts. Likewise, all abbreviations are transcribed as they appear on the page. However, no distinction is made between the letter forms (e.g. a long 'ſ' and a round 's' are transcribed identically as 's' in our diplomatic transcription [Robinson & Solopova, 1993, p. 22]).

|  | Ms. A (Brussels, KBR, 2879-80) | | Ms. B (Brussels, KBR, 2877-78) | |
| --- | --- | --- | --- | --- |
| Train | 1.67% CER | 86/202 folia | 2.36% CER | 107/293 folia |
|  |  | 33,083 words |  | 28,547 words |
| Validation | 2.52% CER | 9/202 folia | 2.81% CER | 11/293 |
|  |  | 3,434 words |  | 2,932 words |

Table 3. Character Error Rates (CERs) for the Handwritten Text Recognition-models trained on large samples of mss. A and B.

## IV SCRIBAL APPROPRIATION

The key question we will deal with in the remainder of this article is the following: how orthographically similar are Hadewijch-manuscripts A and B? Before we present an answer to this question, we will consider how the text is processed (4.1). Next, we will undertake an investigation into which textual elements have the greatest discriminative power to distinguish manuscript A from B (4.2). By definition, these are features that are prevalent in A but not in B, and vice versa (i.e. linguistic idiosyncrasies in which scribal profiles shine through). Finally (4.3), we will consider diachronic trends of variation throughout the manuscripts (i.e. as the text is 'progressing'). Specifically, such an undertaking allows us to ascertain whether there is greater or lesser 'scribal distance' in one part of the manuscripts, compared to other parts. Here, we will pay special attention to the transitions between the textual genres of Hadewijch's oeuvre. Is a scribe inclined to copy more faithfully at the beginning of a new section? Or, do scribal profiles shine through more strongly at the beginning of a text, to then – perhaps – come under the influence of the exemplar, and drift towards its language?

---

[16] https://mufi.info/ [last accessed 06-04-2023].

## 4.1 Preparing the data

Before moving on to the computational analyses, we will briefly elaborate on how the diplomatic transcriptions were preprocessed – albeit this involves only a number of minimal interventions. For one, we opted to consistently restore word breaks. In our original diplomatic transcriptions, words that are split between two lines are in most cases marked by the use of a hyphen. By restoring words that are split between two lines, we can make sure that the analysis is not tainted by circumstances beyond the control of the scribe, such as lack of space on the parchment. Apart from this intervention, all characters specific to the medieval scribal practice (e.g. superscript letters, letters with macrons on top, abbreviation signs such as 'ʒ' or the apostrophe, punctuation, etc.) were retained. Capital letters were also preserved. Their possible height differences, however, were ignored.

As a next step in the pre-processing, the transcriptions from A and B were subdivided into consecutive portions – or: samples. Initially, it was attempted to split the texts at fixed intervals, for example at every 1,000 words as is the case in Kestemont's study [2015, p. 168]. Since the text of A and B runs parallel, one can expect the fifth sample of A to contain a passage similar to the fifth sample of B. For the first ten 1,000-word samples, this is certainly the case. However, after a while, the text in the samples starts to diverge to such an extent that sample A-30 no longer runs parallel to B-30. For this reason, both texts were automatically aligned (at the word-level) before slicing them up into samples. To achieve this, the Needleman-Wunsh sequence alignment algorithm was applied [Needleman & Wunsch, 1970].[17] Next, the produced alignment table (and the aligned text contained therein) was divided into 75 samples (150 samples in total). Each of these samples contains approximately 1,000 words. In short, if we now wish to perform an analysis on, for example, the thirtieth sample, we are confident that these contain similar passages for A and B (see Table 4 for an illustrative example of the produced alignment table).

| A | ī | wilt | ghebruken | . | Eñ | ic | *[...]* | nie | .i. | vre | mi | seluē | bi |
| B | in | wilt | ghebrukē |   | Eñ | ic | *[...]* | nie | een | vre | mi | seluē | bi |

Table 4. Example taken from the alignment table produced for Hadewijch manuscripts A and B. More specifically, this table shows the first and last six aligned tokens of the thirtieth sample.

## 4.2 Distinctive features

In this section, we identify the textual features that have the most salient effect to identify whether a particular sample belongs to manuscript A or B. For this purpose, like Kestemont

---

[17] For our implementation of the Needleman-Wunsch sequence alignment algorithm, we resorted to the wonderful Python library developed by [Folgert Karsdorp, n.d].

[2015, p. 166-168], we first create a bag-of-words model. Similar to 'indexing' in Information Retrieval [Manning et al., 2008, p. 1-18], a bag-of-words model is a compressed representation of one or more documents. To construct such a model, textual elements – most commonly words – are extracted from the documents under scrutiny to then be stored together with their frequencies. Typically, not all words are retained in a bag-of-words model. Rather, a selection is established by means of a frequency cut-off (e.g. only the 100 most frequent words are retained). This reduction of information has statistical advantages, namely: we have a guarantee that our analyses operate on the basis of words observed in all documents [Binongo, 2003, p. 11-12]. As a result, a bag-of-words model is a simplified representation of the text, which can also be rendered as a frequency table. In our case, two bag-of-words models are built, representing the 150 Most Frequent Words (MFWs) in manuscripts A and B. Table 5 shows a limited example of the bag-of-words model created in this way for manuscript A. For each of the 150 MFWs, the model contains accurate information about their absolute frequencies within the samples (A-1, A-2, ... A-75). A similar model was constructed to represent manuscript B.

|  | al | alle | allen | alre | alse | ... | wilt | wāt | w't | zij | zijn |
|---|---|---|---|---|---|---|---|---|---|---|---|
| A-1 | 7 | 6 | 4 | 7 | 8 | ... | 0 | 2 | 1 | 0 | 0 |
| A-2 | 15 | 2 | 0 | 0 | 4 | ... | 3 | 2 | 2 | 0 | 0 |
| A-3 | 7 | 4 | 1 | 0 | 8 | ... | 1 | 6 | 5 | 0 | 0 |
| ... | ... | ... | ... | ... | ... | ... | ... | ... | ... | ... | ... |
| A-74 | 15 | 4 | 7 | 1 | 5 | ... | 0 | 0 | 1 | 0 | 0 |
| A-75 | 16 | 3 | 2 | 0 | 5 | ... | 2 | 0 | 6 | 0 | 0 |

Table 5. Limited portion of the bag-of-words model for the 150 MFWs in 75 samples of manuscript A.

By way of initial exploration, we will probe manuscripts A and B for words with high distinctiveness (also called 'keyness'), using Burrows' Zeta. Originally proposed by John Burrows, Zeta is a method for contrastive textual analysis [Burrows, 2007]. The intuition behind the calculation of Burrows' Zeta is the following: it quantifies the degree of distribution of features (here: words) in two corpora and compares them.[18] Ultimately, Burrows' Zeta returns a list of words that are statistically speaking either preferred or avoided in each subcorpus.

Figure 1 illustrates the result of this initial analysis. In the left half of this figure, words are displayed that are avoided in manuscript A (and thus are characteristic of B). On the right, one can observe words that are preferred in A (and thus are avoided in B). Already from this modest analysis, we can make some interesting observations. First, we notice that the preference is stronger in manuscript B to use abbreviated word forms. In particular, the macron is frequently used. For example, while the expansion of the third-person singular pronoun 'men' ('one') is more prevalent in A, in B the abbreviation 'mē' is more common. The same can be observed for 'den'/'dē' ('the'), 'dan'/'dā' ('than'), 'want'/'wāt' ('because'), and 'allen'/'allē' ('all'). Topping the list as the most preferred word in A is 'hem' ('him').

---

[18] For a formal description of Burrows' Zeta and its variants, we refer the reader to [Schöch et al., 2018].

This is an interesting case, as it – again – demonstrates the predilection in manuscript A for full words instead of abbreviations. It is possible to abbreviate 'hem' as 'hē', which is something the scribe of manuscript B does quite often. However, this creates the risk of confusion, since 'hē' could then in turn either be resolved as 'he*m*' ('him') or 'he*n*' ('they'). In short, by steering away from possibly ambiguous abbreviations, manuscript A lowers the risk of possible misinterpretations of the text.

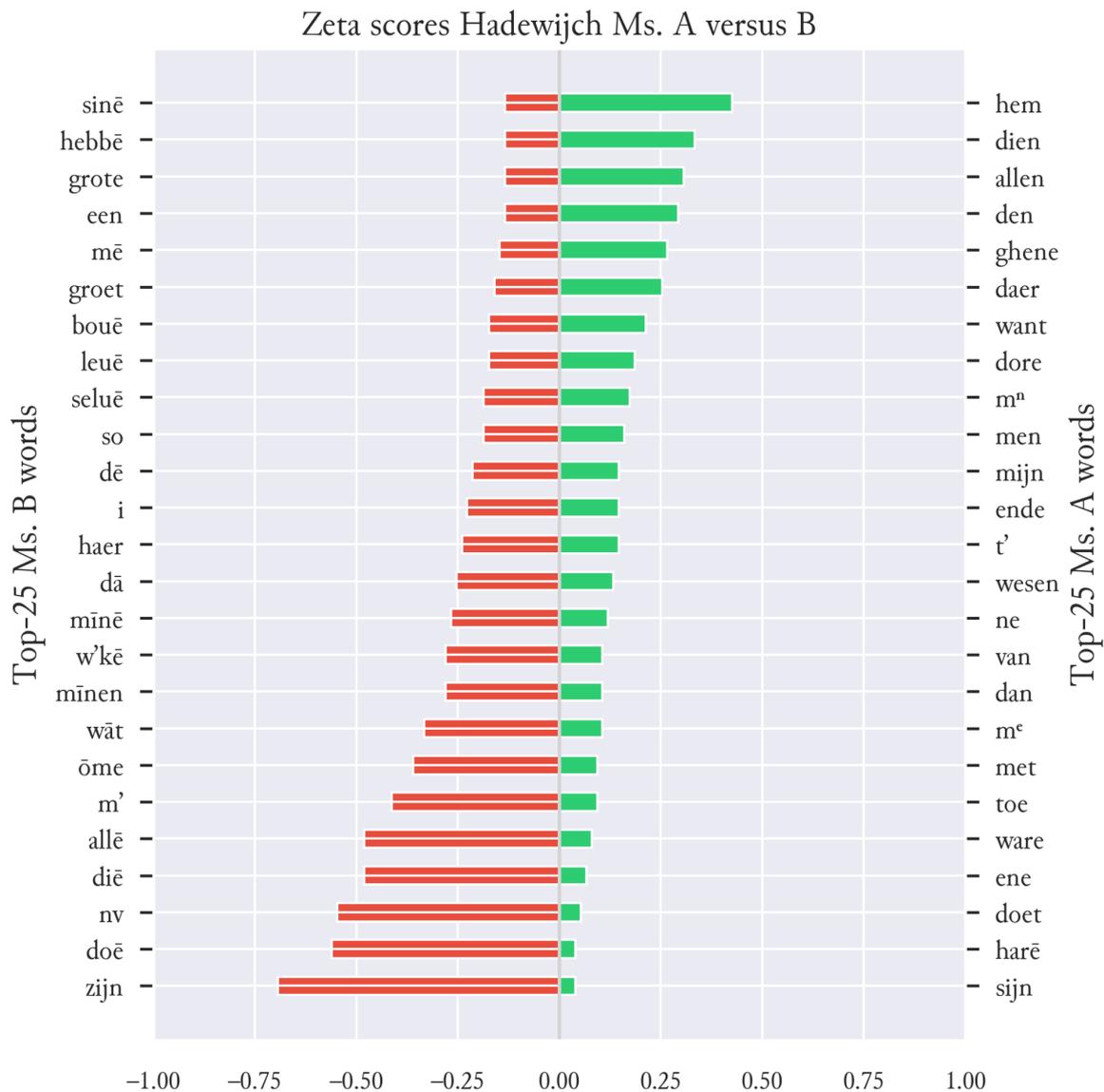

Figure 1. Contrastive textual analysis of Hadewijch-manuscript A versus B, using Burrows' Zeta (on a 1,000-token sample basis).

As a preliminary, exploratory inspection of the distinctive features, Burrows' Zeta offers a useful technique. So far, it allowed us to establish that the scribe of manuscript B appropriated his exemplar at least to some extent. At the same time, however, Burrows' Zeta is sometimes criticised for granting an unfair – and thus in some cases undesirable – advantage to less frequent content words [Schöch et al., 2018]. Therefore, in what follows, we will continue our investigation with a different technique that diminishes the risk of overlooking higher-frequency textual features. To achieve this, we will first replace our bag-of-words model with a different kind of model, containing character-based n-grams (i.e. the result will be a bag-of-n-grams model). Character-based n-grams can be defined as letter sequences of length *n*. For example, the Middle Dutch phrase 'het es' ('it is') can be subdivided into six distinct unigrams (these would be the individual characters), into five bigrams ('he', 'et', 't_', '_e', 'es'), into four trigrams ('het', 'et_', 't_e', '_es'), and so on.[19] The advantage of using character-based n-grams as features for scribal profiling is evident: they allow for a rigorous examination of the spelling preferences of scribes that go beyond atomic word units [Thaisen, 2014; 2020]. Moreover, character n-grams have proven to be very resilient when conducting research on noisy textual data [Eder, 2013]. Even though the HTR-models performed quite well (cf. supra), relying on character n-grams as features is a solid choice. Our newly constructed bag-of-n-grams model represents manuscripts A and B in terms of their 1,000 most frequent character-based 3-grams and 4-grams, thus generously capturing the scribes' spelling behaviour.

For our next experiment, we will rely on a technique from the domain of artificial intelligence, namely the Random Forest Algorithm, which will enable us to determine which n-grams possess the greatest power to distinguish manuscript A from B. The Random Forest Algorithm belongs to a large family of decision tree-models, and is ideally suited for classification tasks [Breiman et al., 1984]. Essentially, it builds a number of tree-like structures – hence: forest – that establish relationships between a random selection of features. In our case, branches are constructed between a random selection of n-grams that might occur in samples of A and B. When a sample enters such a tree, its content is statistically assessed by deciding to what degree an n-gram is present in relation to all the other samples from A and B. To do so, the sample must follow a decision path. Purely as an illustrative example, Figure 2 shows what this process looks like.[20] Let's say the label of a particular 1,000-word sample was forgotten, and we wish to determine whether it more likely stems from manuscript A or B. First, the relative frequency of the 3-gram 'ē w' is analysed. If this value in our anonymous sample is higher than (>) the mean frequency found in most other samples (0.01), the data is split at this point, and we follow the tree to the next node. Here, a decision must be made about the appearance of '_iē' in the sample. If its relative frequency is smaller or equal (≤) than 0.01, the data is split again, and now the occurrence of the 4-gram 'sijn' is evaluated. In this limited example, a final decision now has to be made: if the frequency of 'sijn' is higher than 0.01, the algorithm will classify our anonymous sample as stemming from manuscript A. Otherwise, the verdict goes to B.

---

[19] The underscore ('_') marks a whitespace character.
[20] The visualisation of this decision tree is made possible with the aid of the wonderful Python library `dtreeviz` [Parr & Grover, n.d.].

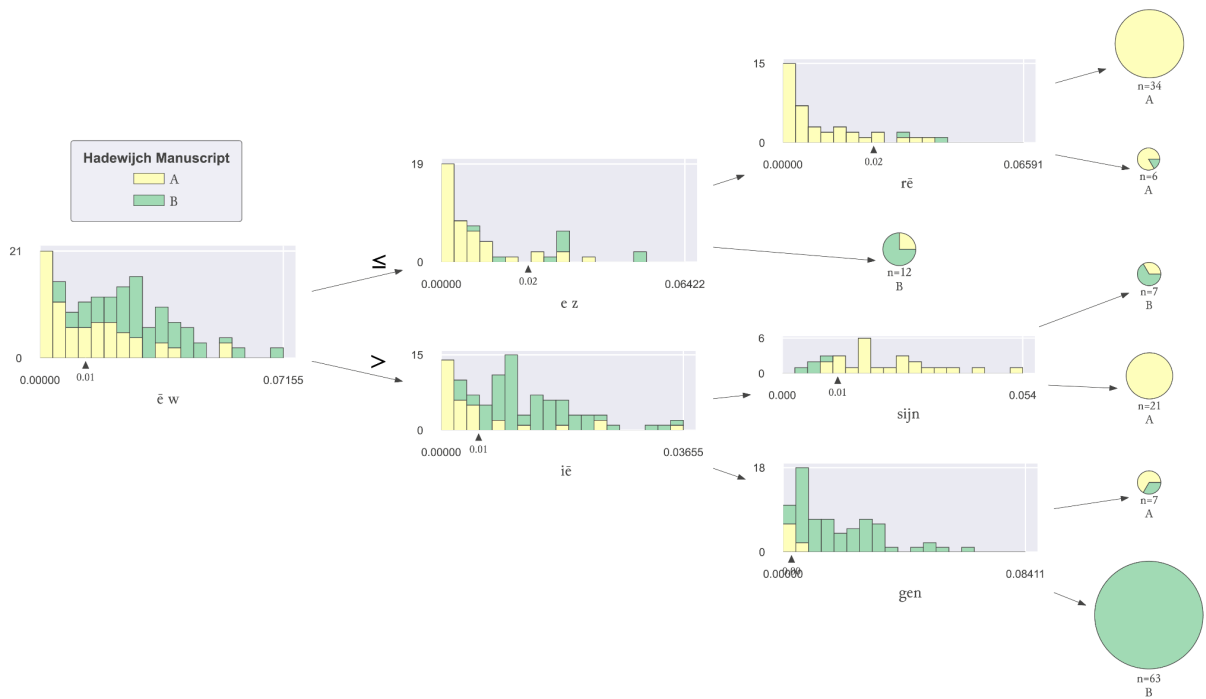

Figure 2. Example of a classification tree generated by the Random Forest algorithm, representing the different selection criteria or 'decision nodes' used to predict the most likely label of a particular manuscript sample.

Random Forest Trees are known to produce both reliable and easily interpretable results. Moreover, what makes them very attractive for scribal profiling research is that they are non-parametric, meaning that they operate without prior knowledge or assumptions about the underlying distribution of the data [Louppe, 2014, p. 25-26]. Another advantage of the Random Forest-method is that it allows for an easy identification of those features that contribute most decisively in reaching an unambiguous, pure classification. An important measure here is Mean Decrease of Impurity (MDI) [Breiman, 2001]. Intuitively, MDI can best be described as a measure for discovering features that are most effective at reaching a clean separation of classes in a tree structure. Hence, in the case of scribal profiling, MDI serves well to identify which n-grams are the driving forces to make accurate predictions.

Figure 3 shows the final result of our experiment with the Random Forest-method. It presents the top-25 n-grams that possess the greatest overall ability to distinguish manuscript A from B. We see that '_zij' has the greatest differentiating power, followed by 'iȷ̄_' and 'zij'. It does not come as a surprise that in particular words spelled with a 'z' account for high distinctiveness. Attention was already drawn to this by the observant Van Mierlo, stating that B contains many more 'z'-spelled words than A [van Mierlo, 1924, vol. 2, p. 5 and 19]. As a feature to differentiate between scribes, however, Kestemont shows that we should not assume an excessively rigid opposition between 's'-spelled and 'z'-spelled words [Kestemont, 2015, p. 172]. Nevertheless, as a feature to distinguish the copy from its exemplar, the opposition 'zijn'-'sijn' seems to be very effective.

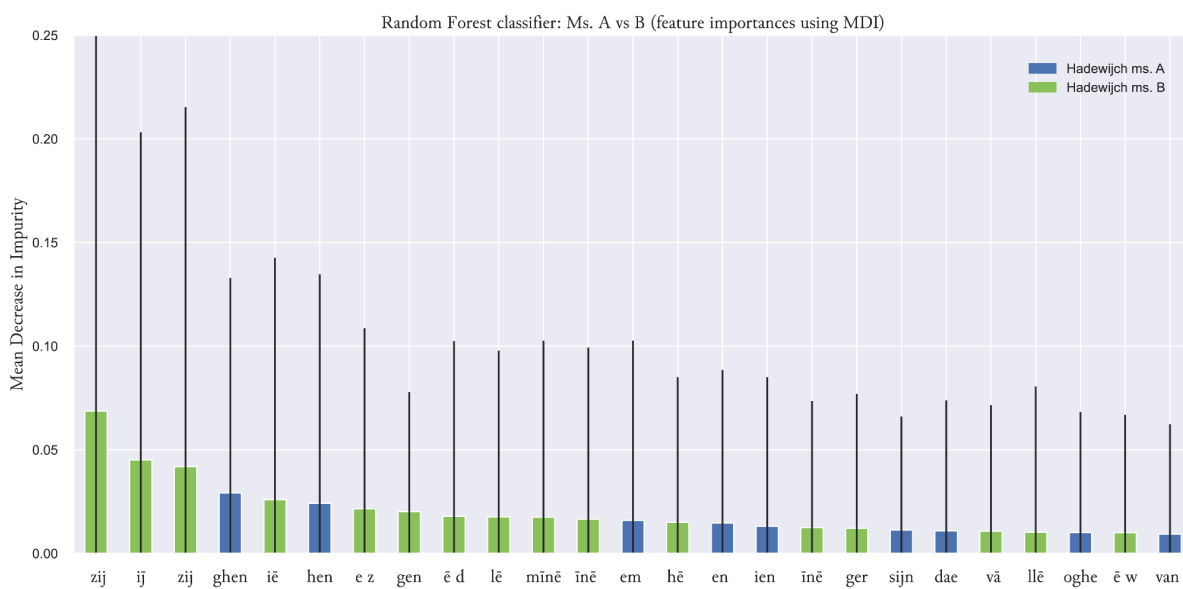

Figure 3. Top-25 of most distinctive character-based 3- and 4-grams for distinguishing between Hadewijch manuscript A and B (relying on Mean Decrease of Impurity).

Next to 'zijn' and 'sijn', Figure 3 also confirms several other observations made by Kestemont [2015]. For example, the n-grams 'ghen', 'gen', and 'ger' prove that there is a clear difference between manuscript A and B when it comes to the insertion of the character 'h'. Contrary to Kestemont's claim, however, the – often superfluous – insertion of 'h' is not a typical feature of B, but of A [ibid., p. 173]. The n-gram 'oghe' demonstrates this as well. Several verb forms, and nouns derived from the verbs 'mogen' ('be able to') and 'dogen' ('to endure') are spelled in B without 'h', whereas in A we notice a preference for the insertion of this character.

Finally, as in Figure 2, we can observe in Figure 3 that 'hen' and 'hē' play an important role in distinguishing between A and B. The scribe of B more frequently uses the abbreviation 'hē' both for 'hem' and 'hen'. This level of efficiency is wasted on A's scribes, who would prefer not to see any confusion between 'God' (to whom the personal pronoun 'hem' usually refers to in Hadewijch's work), and 'they'.

### 4.3 Diachronic trends

As a final experiment, we will look at possible diachronic similarities or discrepancies between manuscripts A and B. Because the different samples of both manuscripts (75 samples for each manuscript) are neatly aligned (i.e. they contain similar passages), we can examine their orthographies in order of appearance. Doing so, we will consider the possibility of three different 'movements', modelled after McIntosh's ideas on dialectical scribal appropriation (cfr. supra; [McIntosh, 1973, p. 61]). For parallel, consecutive portions of an exemplar and its copy, this movement can either be:

1. Consistent – The orthographic profile is consistent when there is about the same amount of variation present in sample *n* as in sample *n*+1.

2. Increasing – When the orthographic profile of sample *n*+1 exhibits more variation than its preceding sample, there is dilation, or an increasing trend of divergence. Specifically, this implies that a scribe's own spelling preferences are incrementally finding their way into his work, to the disadvantage of the language of the exemplar.

3. Narrowing – If the orthographic difference between two consecutive samples becomes smaller, we can observe a narrowing movement, one of closer approximation. The copying scribe comes under the influence, so to speak, of the language of his exemplar and attempts to copy more faithfully.

Figure 4 summarises the result of our investigation into the diachronic trends between exemplar and copy. Much like a seismograph recording the motion of the ground during an earthquake, this figure shows the tremors caused by orthographic variation.[21] The graph should be interpreted as follows: the more horizontal the movement, the more consistent the degree of orthographic variation between exemplar and copy. In the case of a rising slope, the degree of variation increases (i.e. an increasing movement). A negative slope marks a decline in the degree of variation (i.e. a narrowing movement). Moreover, the further the line is removed from the x-axis, the more stark the differences are between samples from A and B.

---

[21] For the setup of the experiment, our bag-of-n-grams model (containing 1,000 features) was sliced up into 500 smaller models of 500 randomly selected features. This resampling technique is also called 'bootstrapping' [Efron & Tibshirani, 1993]. Subsequently, for each of these 500 models (and the features contained therein) it was calculated how they are represented in the individual samples of manuscripts A and B. To this end, the cosine distance was calculated, a distance metric that is commonly used to quantify the stylistic resemblance between literary works [Jannidis et al., 2015]. As a result, for each pair of parallel manuscript samples (i.e. A1-B1, A2-B2, …, A75-B75) 500 similarity scores are obtained for different feature selections. This allows for a more robust inspection of these feature selections (e.g. using percentile statistics for an estimate of the variance).

15Journal of Data Mining and Digital Humanities    http://jdmdh.episciences.org
ISSN 2416-5999, an open-access journal

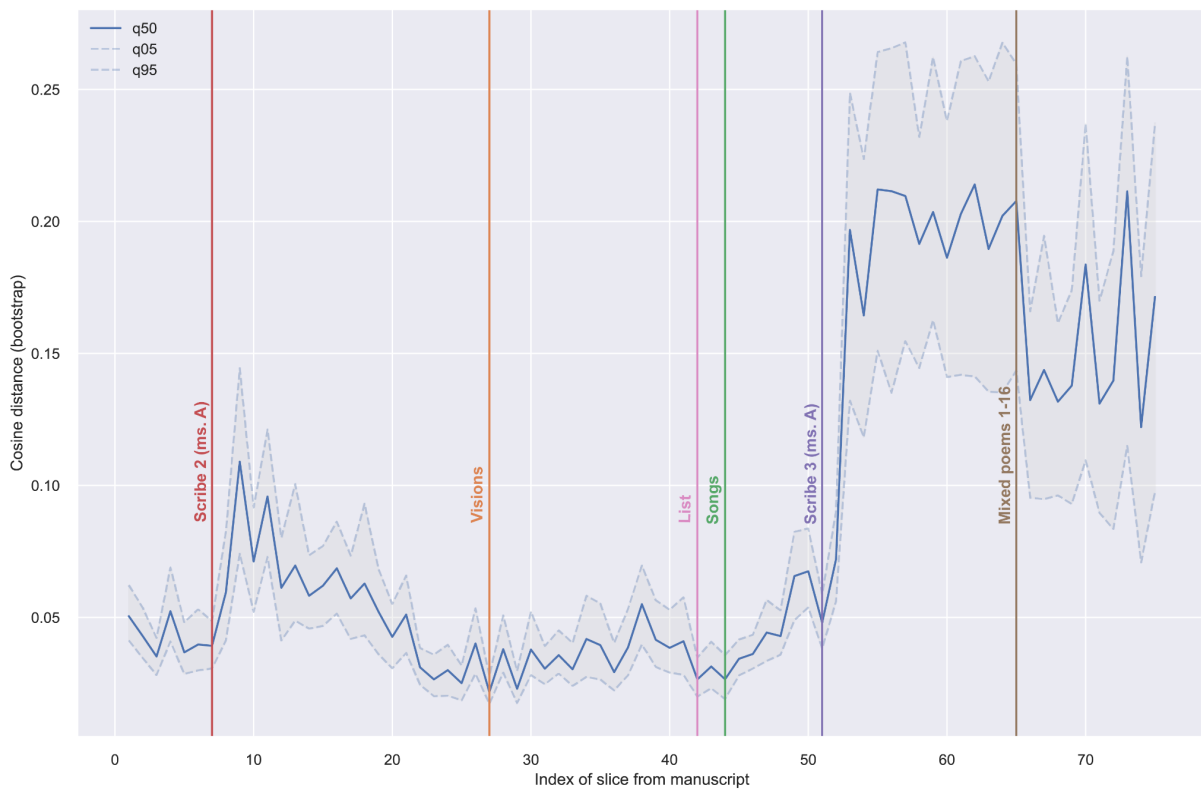

Figure 4. Diachronic trends observed when comparing the orthographies in Hadewijch manuscripts A and B.

First of all, from Figure 4 we learn that the beginning of the manuscript is copied quite faithfully, remaining close to the exemplar's orthography. Starting from the 8th sample (i.e. approximately after 8000 words), a sudden increase can be observed. This coincides roughly with the point at which the second scribe in manuscript A takes over. There are two ways of explaining this. Either the scribe of manuscript B was deliberately cautious at the outset of his copying activity, remaining close to the orthography of his exemplar. Another possibility is that his spelling profile simply closely resembles that of manuscript A's first scribe. It thus remains unclear what the underlying cause is of the similarities between both scribes. However, the transition to the second scribe in manuscript A caused some restraint on the part of manuscript B's scribe. The sudden spike around this transition suggests that he did not readily adopt the spelling of his exemplar at this point. Interestingly, after this peak of orthographic variation, the movement goes down, effectively nearly conforming to the spelling of the exemplar at the beginning of the visions. Also the visions themselves, the *Lijst der volmaakten* and the songs are copied quite faithfully. In this respect, it should be noted that the transition between these different parts of Hadewijch's oeuvre did not prompt the scribe of manuscript B to engage with his exemplar in a different way. What does trigger an effect though, is – once again – a scribal takeover in manuscript A. When the third scribe begins his work in manuscript A, we notice a clear divergence from manuscript B's orthography. Compared to previous sections, the degree of variation increases by a factor of nearly four. Again, this possibly indicates different spelling profiles, and, consequently, a reluctance of manuscript B's scribe to suppress his own spelling preferences in favour of someone else's.

With the start of the mixed poems 1-16, something remarkable happens. The degree of variation significantly drops, and remains fairly consistent throughout four consecutive samples (i.e. over a 4,000-word range, which coincides with poems 1-5). After this – and until the very end of the section containing the mixed poems – the degree of variation fluctuates strongly between the different samples. What is going on here? Several avenues are worth exploring. For example, it could be that transitioning from the songs to the mixed poems, B's scribe felt inclined to follow his exemplar more closely. This, however, seems the least likely explanation, as we saw earlier that such transitions typically do not affect his copying practice. What he is susceptible to, though, are scribal transitions in the exemplar. Then again, we can be relatively certain – on the basis of a palaeographic inspection – that there is no scribal takeover at this point in manuscript A (nor is there one within the mixed poems). Is there perhaps a second scribe at work in manuscript B? This, too, appears to be unlikely, although we can observe that the writing slightly changes at the start of the mixed poems on fol. 132r in manuscript B (Figure 5). Van Mierlo also noted that the writing in B sometimes tends to look differently. Still, both Van Mierlo and Kwakkel conclude that B is the work of a single scribe [van Mierlo, 1924-1925, vol. 2, p. 5; Kwakkel, 1999, p. 39]. We are not inclined to challenge their hypothesis. More likely, the different outlook of the writing on fol. 132r is the result of either a replacement or a sharpening of the pen. This does not dismiss the possibility, however, that starting on fol. 132r, B's scribe recommences his work with renewed energy; perhaps we are witnessing a temporal hiatus between finalising the songs, and starting with the mixed poems 1-16. Finally, and perhaps the most logical answer to the question of what elicits the sudden shift in orthography, we must also consider the possibility of the use of another exemplar. In their seminal work on scribal practices, Reynolds & Wilson note that medieval scribes rarely copied off a single exemplar, but rather tried to compare different copies [Reynold & Wilson, 1991, p. 214]. With some initial restraint, Van Mierlo assumes that a second exemplar was used during the realisation of manuscript B. Supporting evidence is found in a limited number of words and phrases that the scribe of manuscript B could not have witnessed in A [van Mierlo, 1924-1925, vol. 2, p. 12-13]. It is not unimaginable that the scribe of manuscript B had access to another exemplar. As mentioned before, the book list that survives from Rooklooster mentions no less than four copies of Hadewijch's work [cfr. section I]. Possibly, the second exemplar is mentioned on this list, and manuscript B's scribe used it to copy the mixed poems 1-16 from it (and perhaps also the pseudo-Hadewijchiana).

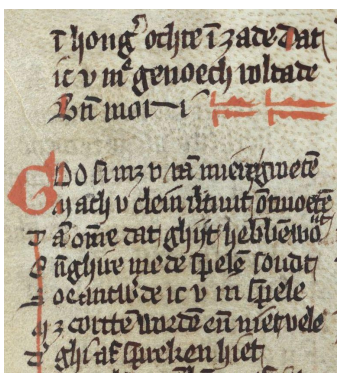

Figure 5. Detail of the start of the mixed poems in Hadewijch ms. B (Brussels, KBR, 2877-78, fol. 132r).

# V CONCLUSION

In this contribution, we pursued a computational analysis of the orthographic profiles found in two of the oldest surviving manuscripts – Brussels, KBR, 2879-2880 (A) and Brussels, KBR, 2877-78 (B) – containing the complete oeuvre of the medieval mystical writer Hadewijch. The foundation of our research was laid by Kestemont [2015], who studied the orthography in Hadewijch's letters, preserved in both manuscripts. The current study considerably extends Kestemont's research by examining the manuscripts from beginning to end, taking into account not only Hadewijch's letters, but also her visions, songs and mixed poems.

Arriving at the end of our examination, we can summarise the main findings into two groups, relating either to the methodology, or to the concrete implications for the Hadewijch-scholarship. Firstly, by means of HTR-technology, one can nowadays unlock a wealth of handwritten material with relative ease. Of course, automatically obtained transcriptions do not deserve the same hallmark as manually transcribed, carefully checked diplomatic transcriptions. However, for research purposes such as scribal profiling, they prove more than adequate. The usefulness of HTR-transcriptions is further advanced by our choice of features and the technology employed to probe them. In line with Thaisen's methodology [2014; 2020], we observed that character-based n-gram models are particularly well-suited for investigating scribal copying practices. Moreover, they have been demonstrated to be fairly resistant to influences stemming from HTR-related transcription errors [Eder 2013]. Techniques from the field of artificial intelligence proved particularly useful in our search for the most distinctive features of both manuscripts. In particular, the Random Forest-algorithm is a useful method, thanks to its pragmatic approach to the data, and the interpretable results it yields.

Secondly, relating to the Hadewijch-manuscripts under scrutiny, we were able to establish that the scribe of manuscript B appropriated the orthography of his exemplar in various ways. Prior observations made by Kestemont [2015] remain valid (except that 'h'-insertion in words like 'heilighen' and 'ghenoech' is more characteristic of A than of B). Importantly, the fact that linguistic scribal appropriation can be detected and quantified is in itself an interesting result, given that both A and B are written in the same vernacular (i.e. Brabantian, a dialect of the Middle Dutch language) [van Mierlo, 1924-1925, vol. II, p. 21]. Thus, we can not only perform inter-dialectic scribal profiling (as is the case in McIntosh' original conceptualisation of the 'linguistic scribal profile'), but also *within* the same dialect, we are able to pinpoint the idiosyncratic traits of different scribes. Another important contribution we have been able to achieve relates to the identification of possible diachronic scribal trends. Among other things, we could observe that the orthographic effect of scribal takeovers in manuscript A percolates into its copy. Finally, we also examined whether transitions between different content matters had an impact on the copying practice of B's scribe. This was found not to be the case. However, when the scribe in manuscript B arrives at the last part of Hadewijch's oeuvre – the mixed poems 1-16 – we notice a sudden decrease in orthographic variation. We could not provide a conclusive answer as to what spurs this sudden shift, yet we wish to put forward the tentative hypothesis that this part was copied not off manuscript A, but off another exemplar. This latter finding deserves further investigation;

which, in our view, can only be done by broadening our perspective, and taking into account all manuscripts containing Hadewijch's oeuvre.

# IV REFERENCES AND CITATIONS

Alcorn, R., Kopaczyk, J., Los, B., & Molineaux, B. (2019). Historical Dialectology and the Angus McIntosh Legacy. In R. Alcorn, J. Kopaczyk, B. Los, & B. Molineaux, *Historical Dialectology in the Digital Age* (pp. 1–16). Edinburgh University Press.

Andrews, T. L., & Mace, C. (2013). Beyond the tree of texts: Building an empirical model of scribal variation through graph analysis of texts and stemmata. *Literary and Linguistic Computing*, *28*(4), 504–521.

Arblaster, J., & Faesen, R. (2014). Mysticism in the Low Countries before Ruusbroec. In J. Arblaster & R. Faesen (eds.), *A companion to John of Ruusbroec* (pp. 5–46). Brill.

Axters, S. G. (1953). *Geschiedenis van de vroomheid in de Nederlanden. II. De eeuw van Ruusbroec*. De Sikkel.

Axters, S. G. (1964). Hadewijch als voorloopster van de zalige Jan van Ruusbroec. In *Dr. L. Reypens-Album. Opstellen aangeboden aan Prof. Dr. L. Reypens s.j. Ter gelegenheid van zijn tachtigste verjaardag op 26 februari 1964* (pp. 57–74). Ruusbroecgenootschap.

Binongo, J. (2003). Who wrote the 15th book of Oz? An application of multivariate analysis to authorship attribution. *Chance*, *16*(2), 9–17.

Black, M. (1998). A scribal translation of 'Piers Plowman'. *Medium Ævum*, *67*(2), 257–290.

Boot, P., & Brinkman, H. (n.d.). *Richtlijnen voor digitale diplomatische edities in de reeks 'Middeleeuwse Verzamelhandschriften uit de Nederlanden'*. Retrieved 11 October 2022, from https://github.com/HuygensING/mvn-xml/blob/main/docu/Richtlijnen%20MVN%20digitaal.pdf

Breiman, L. (2001). Random Forests. *Machine Learning*, *45*(1), 5–32.

Breiman, L., Friedman, J., Olshen, R., & Stone, C. (1984). *Classification and regression trees*. Chapman & Hall/CRC.

Burrows, J. (2007). All the way through. Testing for authorship in different frequency strata. *Literary and Linguistic Computing*, *22*(1), 27–47.

Carrillo-Linares, M. J. (2020). Copying strategies of late Middle English scribes: Hand(s) and language(s) of two 15th-century manuscripts. *SELIM. Journal of the Spanish Society for Medieval English Language and Literature*, *25*(1), 121–172.

Deschamps, J. (1970). *Middelnederlandse handschriften uit Europese en Amerikaanse bibliotheken*.

Driscoll, M. J. (2010). The words on the page: Thoughts on philology, old and new. In J. Quinn & E. Lethbridge (eds.), *Creating the medieval saga: Versions, variability, and editorial interpretations of Old Norse saga literature.* (pp. 85–102). Syddansk Universitetsforlag.

Eder, M. (2013). Mind your corpus: Systematic errors in authorship attribution. *Literary and Linguistic Computing*, *28*(4), 603–614.

Efron, B., & Tibshirani, R. (1993). *An introduction to the bootstrap*. Chapman & Hall.

Fraeters, V. (2013). Hadewijch of Brabant and the Beguine Movement. In E. Andersen, H. Lähnemann, & A. Simon (eds.), *A companion to mysticism and devotion in northern Germany in the late middle ages* (pp. 49–71). Brill.

Fraeters, V., Willaert, F., & Grijp, L. P. (eds.). (2009). *Hadewijch. Liederen*. Historische Uitgeverij.

Hart, C. (1980). *Hadewijch. The complete works*. Paulist Press.

Hiltunen, R., & Peikola, M. (2007). Trial discourse and manuscript context: Scribal profiles in the Salem witchcraft records. *Journal of Historical Pragmatics*, *8*(1), 43–68.

Jannidis, F., Pielström, S., Schöch, C., & Vitt, T. (2015). Improving Burrows' Delta – An empirical evaluation of text distance measures. *Digital Humanities 2015 Conference Abstracts*. Digital Humanities 2015, Sydney.

Kahle, P., Colutto, S., Hackl, G., & Mühlberger, G. (2017). Transkribus—A Service Platform for Transcription, Recognition and Retrieval of Historical Documents. *14th IAPR International Conference on Document Analysis and Recognition (ICDAR)*, *04*, 19–24.

Karsdorp, F. (n.d.). *(Multiple) Sequence Alignment*. Retrieved 8 October 2022, from https://github.com/fbkarsdorp/alignment

Kestemont, M. (2015). A computational analysis of the scribal profiles in two of the oldest manuscripts of Hadewijch's letters. *Scriptorium*, *69*, 159–177.

Kiessling, B., Tissot, R., Stokes, P., & Stökl Ben Ezra, D. (2019). eScriptorium: An Open Source Platform for Historical Document Analysis. *International Conference on Document Analysis and Recognition Workshops (ICDARW)*, *2*, 19–24.

Kwakkel, E. (1999). 'Ouderdom en genese van de veertiende-eeuwse Hadewijch-handschriften'. *Queeste*, *6*, 23–40.

Kwakkel, E. (2002). *Die dietsche boeke die ons toebehoeren: De kartuizers van Herne en de productie van Middelnederlandse handschriften in de regio Brussel (1350-1400)*. Peeters.

Lievens, R. (1971). 'Toebehoren', 'hebben' en de lijst der Dietse boeken van Rooklooster. *Tijdschrift Voor Nederlandse Taal- En Letterkunde*, *87*, 156-160.

Louppe, G. (2014). *Understanding Random Forests. From theory to practice*. University of Liège. arXiv:1407.7502

Manning, C. D., Raghavan, P., & Schütze, H. (2008). *Introduction to information retrieval*. Cambridge University Press.

McIntosh, A. (1973). Word geography in the lexicography of medieval English. *Annals of the New York Academy of Sciences*, *211*(1), 55–66.

McIntosh, A. (1974). Towards an inventory of Middle English scribes. *Neuphilologische Mitteilungen*, *75*(4), 602–624.

McIntosh, A. (1975). Scribal profiles from Middle English texts. *Neuphilologische Mitteilungen*, 218–235.

Mommaers, P. (2003). *Hadewijch. Schrijfster—Begijn—Mystica*. Peeters.

Muehlberger, G., Seaward, L., Terras, M., Ares Oliveira, S., Bosch, V., Bryan, M., Colutto, S., Déjean, H., Diem, M., Fiel, S., Gatos, B., Greinoecker, A., Grüning, T., Hackl, G., Haukkovaara, V., Heyer, G., Hirvonen, L., Hodel, T., Jokinen, M., … Zagoris, K. (2019). Transforming scholarship in the archives through handwritten text recognition: Transkribus as a case study. *Journal of Documentation*, *75*(5), 954–976.

Needleman, S. B., & Wunsch, C. D. (1970). A general method applicable to the search for similarities in the amino acid sequence of two proteins. *Journal of Molecular Biology*, *48*(3), 443–453.

Parr, T., & Grover, P. (n.d.). *dtreeviz: Decision tree visualization*. Retrieved 8 October 2022, from https://github.com/parrt/dtreeviz

Reynaert, J. (1981). Ruusbroec en Hadewijch. *Ons Geestelijk Erf*, *55*(3), 193–233.

Robinson, P., & Solopova, E. (1993). Guidelines for transcription of the manuscripts of the Wife of Bath's Prologue. *The Canterbury Tales Project. Occasional Papers*, *I*.

Ruh, K. (1992). Fragen und Beobachtungen zu den Poetica der Hadewijch. In J. Janota, P. Sappler, F. Schanze, B. K. Vollmann, G. Vollmann-Profe, & H.-J. Ziegeler (eds.), *Festschrift Walter Haug und Burghart Wachinger* (pp. 323–334). Max Niemeyer Verlag.

Schöch, C., Schlör, D., Zehe, A., Gebhard, H., Becker, M., & Hotho, A. (2018). Burrows' Zeta: Exploring and evaluating variants and parameters. In J. G. Palau & I. G. Russell (eds.), *Digital Humanities* (pp. 274–277).

Thaisen, J. (2014). Initial position in the Middle English verse line. *English Studies*, *95*(5), 500–513.

Thaisen, J. (2020). Standardisation, exemplars, and the Auchinleck manuscript. In L. Wright (ed.), *The Multilingual Origins of Standard English* (Vol. 9, pp. 165–190).

van Mierlo, J. (ed.). (1924). *Hadewijch. De visioenen* (Vol. 1–2). Vlaamsche Boekenhalle.

van Mierlo, J. (1927). Beata Hadewigis de Antverpia. *Dietsche Warande en Belfort*, *27*, 787–798, 833–843.

van Mierlo, J. (ed.). (1942). *Hadewijch. Strophische gedichten* (Vol. 1–2 delen). Standaard-Boekhandel.

van Mierlo, J. (ed.). (1952). *Hadewijch. Mengeldichten*. Standaard Boekhandel.

Verheyden, P. (1937). *Huis en have van Godevaert de Bloc, scriptor en boekbinder te Brussel, 1364-1384*. Martinus Nijhoff.

Willaert, F. (2013). Dwaalwegen. Recente hypotheses over Hadewijchs biografie. *Ons Geestelijk Erf 84*, *2*, 153–194.